\documentclass{llncs}
\usepackage[utf8]{inputenc}
\usepackage{multirow}
\usepackage{amsmath}
\usepackage{amssymb} 
\usepackage[nolist,nohyperlinks]{acronym}
\usepackage{url}
\usepackage{arydshln}

\title{A Multilingual Study of Compressive Cross-Language Text Summarization}

\author{Elvys Linhares Pontes\inst{1,2,3} \and Stéphane Huet\inst{1} \and Juan-Manuel Torres-Moreno\inst{1,2,3}}
\authorrunning{E. Linhares Pontes et al.}

\institute{LIA, Université d'Avignon et des Pays de Vaucluse, Avignon, 84000 France\\
\email{elvys.linhares-pontes@alumni.univ-avignon.fr}
\and D\'epartement de GIGL, \'Ecole Polytechnique de Montr\'eal,\\
C.P. 6079, succ. Centre-ville, Montr\'eal (Qu\'ebec) H3C 3A7 Canada
\and Laboratoire GDAC, Université du Québec à Montréal,\\
C.P. 8888, succ. Centre-ville
Montréal (Québec) H3C 3P8 Canada}

\begin{document}

\maketitle

\begin{abstract}
Cross-Language Text Summarization generates summaries in a language different from the language of the source documents. 
Recent methods use information from both languages to generate summaries with the most informative sentences.
However, these methods have performance that can vary according to languages, which can reduce the quality of summaries. 
In this paper, we propose a compressive framework to generate cross-language summaries.
In order to analyze performance and especially stability, we tested our system and extractive baselines on a dataset available in four languages (English, French, Portuguese, and Spanish) to generate English and French summaries.
An automatic evaluation showed that our method outperformed extractive state-of-art CLTS methods with better and more stable ROUGE scores for all languages.
\end{abstract}

\section{Introduction}

Cross-Language Text Summarization (CLTS) aims to generate a summary of a document, where the summary language differs from the document language.
Many of the state-of-the-art methods for CLTS are of the extractive class. 
They mainly differ on how they compute sentence similarities and alleviate the risk that translation errors are introduced in the produced summary. 

Previous works analyze the CLTS only between two languages for a given dataset, which does not demonstrate the stability of methods for different texts and languages.
Recent works carried out compressive approaches based on neural networks, phrase segmentation, and graph theory~\cite{Yao:2015:phrase_based,Wan:2018,elvys:2018:nldb}.
Among these models, Linhares Pontes \textit{et al.} introduced the use of chunks and two compression methods at the sentence and multi-sentence levels to improve the informativeness of cross-language summaries~\cite{elvys:2018:nldb}.

In this paper, we adapt the method presented in \cite{elvys:2018:nldb} to perform CLTS for several languages.
More precisely, we modified the creation of chunks and we simplified their multi-sentence compression method to be able to analyze several languages and 
compress small clusters of similar sentences.
To demonstrate the stability of our system, we extend the MultiLing Pilot dataset~\cite{Giannakopoulos:2011} with two Romance languages (Portuguese and Spanish) to test our system to generate \{French, Portuguese, Spanish\}-to-English and \{English, Portuguese, Spanish\}-to-French cross-language summaries.
Finally, we carried out an automatic evaluation to make a systematic performance analysis of systems, which details the characteristics of each language and their impacts on the cross-language summaries.

The remainder of this paper is organized as follows. 
Section~\ref{sc:cclts} details our contributions to the compressive CLTS approach.
In Section~\ref{sc:rw}, we describe the most recent works about CLTS.
Section~\ref{sc:exp} reports the results achieved on the extended version of the MultiLing 2011 dataset and the analysis of cross-language summaries.
Finally, conclusions and future work are set out in Section~\ref{sc:conc}.

\section{Compressive Cross-Language Text Summarization}
\label{sc:cclts}

Following the approach proposed by Linhares Pontes \textit{et al.} \cite{elvys:2018:nldb}, we combined the analysis of documents in the source and the target languages, with Multi-Sentence Compression (MSC) to generate more informative summaries. 
We expanded this approach in three ways.

In order to simplify and to extend the analysis to several languages, we only use Multi-Word Expressions for the English target language and we replace the analysis of parallel phrases with syntactic patterns to create chunks. 
Then, we also optimized the MSC method for small clusters by removing the analysis of 3-grams.
Unfortunately, we have not found any available dataset for sentence compression in other languages; therefore, we restrict the use of compressive methods to MSC.
Finally, differently from \cite{elvys:2018:nldb}, compressed versions of sentences were considered in the CoRank method instead of the only original versions, in order to estimate the relevance of sentences for summaries. 

The following subsections highlight our contributions to the architecture of the method presented in \cite{elvys:2018:nldb}.

\subsection{Preprocessing}

Initially, source texts are translated into English and French with the Google Translate system\footnote{\url{https://translate.google.com}}, which was used in the majority of the state-of-the-art CLTS methods \cite{Wan:2011,Wan:2018,elvys:2018:nldb}.

Then, a chunk-level tokenization is performed on the target language side \cite{Villada:2006,deCaseli:2010}.
We applied two simple syntactic patterns to identify useful structures: $<(ADJ)^*(NP|NC)^+>$ for English and $<(ADJ)^*(NP|NC)^+(ADJ)^*>$ for French, where ADJ stands for adjective, NP for proper noun and NC for common noun.
We also use the Stanford CoreNLP tool \cite{StanfordCoreNLP14} for the English translations. This tool detects phrasal verbs, proper names, idioms and so on.
Unfortunately, we did not find a similar tool for French; consequently, the French chunk-level tokenization is limited to the syntactic pattern.

\subsection{Multi-Sentence Compression}




We aim to generate a single, short, and informative compression from clusters of similar sentences. Therefore, we use the Linhares Pontes \textit{et al.}'s method \cite{elvys:2018:nldb} to create clusters of similar sentences based on their similarity in the source and the target languages.

As the majority of clusters are composed of few similar sentences (normally two or three sentences), 3-grams are not frequent and the associated score is of little interest for the compress process.
Therefore, we simplify the Linhares Pontes \textit{et al.}'s method \cite{elvys:2018:nldb,linhares:2018:textgraph} to process MSC guided only by the cohesion of words and keywords.
%
%
%
%

Our MSC method looks for a sentence that has a good cohesion and the maximum  of keywords, inside a word graph built for each cluster of similar sentences according to the method devised by Filippova~\cite{filippova:2010}. In this graph, arcs between two vertices representing words (or chunks) are weighted by a cohesion score that is defined by the frequency of these words inside the cluster. Vertices are labeled depending on whether they are or not a keyword identified by the Latent Dirichlet Allocation (LDA) method inside the cluster (see \cite{elvys:2018:nldb} for more details). From these scores and labels, the MSC problem is expressed as the following objective:

\begin{equation}
  \mathrm{minimize} ~ 
	\Big( \sum_{(i,j) \in A} w(i,j) \cdot x_{i,j}  - k \cdot \sum_{l \in L} b_l
    \Big) \label{fo}
\end{equation}

\noindent where $x_{ij}$ indicates the existence of the arc $(i,j)$ in the solution, $w(i,j)$ is the cohesion of the words $i$ and $j$, 
$L$ is the set of labels (each representing a keyword), $b_l$ indicates the existence of a chunk with a keyword $l$ in the solution, $k$ is the keyword bonus of the graph\footnote{The keyword bonus is defined by the geometric average of all weight arcs in the graph and aims at favoring compressions with several keywords.}.
Finally, we generate the 50 best solutions according to the objective (\ref{fo}) and we select the compression with the lowest normalized score (Equation \ref{eq:snorm}) as the best compression:

\begin{equation} \label{eq:snorm}
   \textrm{score}(c) = \frac{e^{\textrm{opt}(c)} }{||c||},
\end{equation}

\noindent where $\textrm{opt}(c)$ is the score of the compression $c$ from Equation \ref{fo}. We restrict the MSC method to the sentences in the target language in order to avoid errors generated by machine translation, which would be applied in a post-processing step on compressed sentences.

\subsection{CoRank Method}
\label{ssc:corank}

Sentences are scored based on their information in both languages using the CoRank method \cite{Wan:2011} which analyzes sentences in each language separately, but also between languages (Equations \ref{eq:corank:begin}--\ref{eq:corank:end}). 

\begin{equation}
\label{eq:corank:begin}
\mathbf{u} = \alpha \cdot (\mathbf{\tilde{M}^{sc}})^T \mathbf{u} + \beta \cdot (\mathbf{\tilde{M}^{tg,sc}})^T \mathbf{v}
\end{equation}
\begin{equation}
\mathbf{v} = \alpha \cdot (\mathbf{\tilde{M}^{tg}})^T \mathbf{v} + \beta \cdot (\mathbf{\tilde{M}^{tg,sc}})^T \mathbf{u}
\end{equation}
\begin{equation}
M_{ij}^\textrm{tg} = 
\begin{cases}
    \textrm{cosine}(s_i^{tg}, s_j^{tg}), &  \textrm{if}\ i \neq j \\
    0 & \text{otherwise}
  \end{cases}
\end{equation}
\begin{equation}
M_{ij}^\textrm{sc} =
  \begin{cases}
    \textrm{cosine}(s_i^{sc}, s_j^{sc}), & \textrm{if}\ i \neq j \\
    0 & \text{otherwise}
  \end{cases}
\end{equation}
\begin{equation}
\label{eq:corank:end}
\textrm{M}_{ij}^\textrm{tg,sc} = \sqrt{\textrm{cosine}(s_i^{sc}, s_j^{sc}) \times \textrm{cosine}(s_i^{tg}, s_j^{tg})}
\end{equation}

\noindent where $\textrm{M}^\textrm{tg}$ and $\textrm{M}^\textrm{sc}$ are normalized to $\tilde{\textrm{M}}^\textrm{tg}$ and $\tilde{\textrm{M}}^\textrm{sc}$, respectively, to make the sum of each row equal to 1. 
$\mathbf{u}$ and $\mathbf{v}$ denote the relevance of the source and target language sentences, respectively. $\alpha$ and $\beta$ specify the relative contributions to the final scores from the information in the source and the target languages, with $\alpha + \beta = 1$.

Finally, summaries are generated with the most relevant sentences in the target language. We add a sentence/compression to the summary only if it is sufficiently different from the sentences/compressions already in the summary.

\section{Related Work}
\label{sc:rw}

Cross-Language Text Summarization schemes can be divided in early and late translations, and joint analysis.
The early translation first translates documents to the target language, then it summarizes these translated documents using only information of these translations. 
The late translation scheme does the reverse.
The joint analysis combines the information from both languages to extract the most relevant information.

Regarding the analysis of machine translation quality, Wan \textit{et al.} \cite{Wan:2010} and Boudin \textit{et al.} \cite{boudin:2011} used sentence features (sentence length, number of punctuation marks, number of phrases in the parse tree) to estimate the translation quality of a sentence. Wan \textit{et al.} used an annotated dataset made of pairs of English-Chinese sentences with translation quality scores to train their Support Vector Machine (SVM) regression method. Finally, sentences that have a high translation quality and a high informativeness were selected for the summaries. Similarly, Boudin \textit{et al.} trained an $e$-SVR using a dataset composed of English and automatic French translation sentences to calculate the translation quality based on the NIST metrics. Then, they used the PageRank algorithm to estimate the relevance of sentences based on their similarities and translation quality.
Yao \textit{et al.} devised a phrase-based model to jointly carry out sentence scoring and sentence compression \cite{Yao:2015:phrase_based}.
They developed a scoring scheme for the CLTS task based on a submodular term of compressed sentences and a bounded distortion penalty term.

Wan \cite{Wan:2011} leverages the information in the source and in the target language for cross-language summarization. 
He proposed two graph-based summarization methods (SimFusion and CoRank) for the English-to-Chinese CLTS task. 
The first method linearly fuses the English-side and Chinese-side similarities for measuring Chinese sentence similarity.
In a nutshell, this method adapts the PageRank algorithm to calculate the relevance of sentences, where the weight arcs are obtained by the linear combination of the cosine similarity of pairs of sentences for each language. 
The CoRank method was described in Section \ref{ssc:corank}.

Recently, Wan \textit{et al.} \cite{Wan:2018} carried out the cross-language document summarization task by extraction and ranking of multiple summaries in the target language.
They analyzed many summaries in order to produce high-quality summaries for every kind of documents.
Their method uses a top-K ensemble ranking for candidate summary, based on features that characterize the quality of a candidate summary.
They used multiple text summarization and machine translation methods to generate the summaries.

In order to generate abstractive cross-lingual summaries, Zhang \textit{et al.} \cite{Zhang:2016} extended the work of Bing \textit{et al.} \cite{Bing:2015} that constructs new sentences by exploring noun/verb phrases. Their method first constructs a pool of concepts and facts represented by phrases in English and Chinese translation sentences. Then new sentences are generated by selecting and merging informative phrases in both languages to maximize the salience of phrases and meanwhile satisfy the sentence construction constraints. They employ integer linear optimization for conducting phrase selection and merging simultaneously in order to generate informative cross-lingual summaries with a good translation quality.
This method generates abstractive summaries; however, the framework to identify concepts and facts only works for English, which prevents this method from being extended for other languages.

\section{Experimental Evaluation}
\label{sc:exp}

We estimate the performance of our approach in relation to the early and the late translations, SimFusion and CoRank methods\footnote{We used the same configuration for SimFusion and CoRank as described in \cite{elvys:2018:nldb}.}\footnote{Unfortunately, the majority of state-of-the-art systems in CLTS are not available. Therefore, we only considered extractive systems in our analysis.}.
All systems generate summaries containing a maximum of 250 words with the best scored sentences without redundant sentences.
We regard a similarity score (cosine similarity) with a threshold $\theta$ of 0.6 to create clusters of similar sentences for the MSC\footnote{We use the same threshold $\theta$ of 0.5 described in \cite{elvys:2018:nldb} for French-to-English cross-language summaries.} and a threshold $\theta$ of 0.5 to remove redundant sentences in the summary generation.

\subsection{Datasets}

We used the English and French language versions of the MultiLing Pilot 2011 dataset \cite{Giannakopoulos:2011}.
This dataset contains 10 topics which have 10 source texts and 3 reference summaries per topic.
These summaries are composed of 250 words.
In order to extend the analysis to other languages, English source texts were translated into the Portuguese and Spanish languages by native speakers \footnote{The extension of the MultiLing Pilot 2011 dataset is available at: \url{http://dev.termwatch.es/~fresa/CORPUS/TS/}}.
Specifically, we use English, French, Portuguese, and Spanish texts to test our system.

\subsection{Evaluation}

An automatic evaluation with ROUGE \cite{lin:2004} was carried out to compare the differences between the distribution of $n$-grams of the candidate summary and a set of reference summaries.
More specifically, we used unigram (ROUGE-1 or R-1), bigram (ROUGE or R-2), and skip-gram (ROUGE-SU4 or R-SU4) analyses.

Table \ref{tb:clts:french} describes the ROUGE f-scores obtained by each system to generate French summaries from English, Portuguese, and Spanish source texts.
Despite using the information from both languages, the SimFusion method achieved comparable results with respect to the early and late approaches.
On the contrary, CoRank and our approach consistently obtained better results than other baselines, with at least an absolute difference of 0.035 in ROUGE-1 for all languages.
The MSC method improved the CoRank method by generating more informative compressions for all languages.
The last two lines show that chunks helped our MSC method to generate slightly more informative summaries (better ROUGE-1 scores).

\begin{table}[h]
\centering
\caption{\label{tb:clts:french}ROUGE f-scores for cross-language summaries from English, Portuguese, and Spanish languages to French language.}
\begin{tabular}{l|ccc|ccc|ccc}
\hline
\multirow{2}{*}{Methods}       & \multicolumn{3}{c|}{English}  & \multicolumn{3}{c|}{Portuguese}  & \multicolumn{3}{c}{Spanish}  \\
         & R-1      & R-2      & R-SU4    & R-1      & R-2      & R-SU4    & R-1      & R-2      & R-SU4    \\ \hline \hline
\rule{0pt}{10pt}Late           &  0.4190 & 0.0965 & 0.1588 & 0.4403 & 0.1128 & 0.1746 & 0.4371 & 0.1133 & 0.1738 \\ \hline
\rule{0pt}{10pt}Early          &  0.4223 & 0.1007 & 0.1631 & 0.4386 & 0.1110 & 0.1743 & 0.4363 & 0.1143 & 0.1729 \\ \hline
\rule{0pt}{10pt}SimFusion      &  0.4240 & 0.1004 & 0.1637 & 0.4368 & 0.1105 & 0.1735 & 0.4350 & 0.1125 & 0.1723 \\ \hline
\rule{0pt}{10pt}CoRank         &  0.4733 & 0.1379 & 0.1963 & 0.4723 & 0.1460 & 0.2006 & 0.4713 & 0.1387 & 0.1942 \\  \hline
\rule{0pt}{10pt}Our approach   &  \textbf{0.4831} & 0.1460 & \textbf{0.2030} & \textbf{0.4784} & 0.1511 & \textbf{0.2045} & \textbf{0.4825} & 0.1481 & 0.2050 \\ \hline
\rule{0pt}{10pt}\begin{tabular}[c]
{@{}l@{}}Our approach\\w/o chunks
\end{tabular}                  &  0.4817 & \textbf{0.1463} & 0.2021 & \textbf{0.4784} & \textbf{0.1518} & 0.2044 & 0.4805 & \textbf{0.1486} & \textbf{0.2056} \\ \hline
\end{tabular}
\end{table}

The Multiling dataset is composed of 10 topics in several languages; however, these topics are expressed in different ways for each language. These dissimilarities implies a variety of vocabulary sizes and sentence lengths, and, consequently, of outputs of the MT system from each source language (Table~\ref{tb:datasets:french}). The biggest difference in the statistics is between English source texts and its French translation vocabulary. French translations significantly increased the vocabulary from English source texts and the number of words. These translations also are longer than source texts, except for the Spanish that has similar characteristics. 
Our simple syntactic pattern created similar numbers of chunks for all languages with the same average length.
The addition of these simple chunks did not significantly improve the informativeness of our compressions. 

\begin{table}[h]
\centering
\caption{\label{tb:datasets:french}Statistics of datasets and their translation to French.}
\begin{tabular}{l|cc|cc|cc}
\hline
                & \multicolumn{2}{c|}{English} & \multicolumn{2}{c|}{Portuguese} & \multicolumn{2}{c}{Spanish} \\
                & Source    & Fr-Translation   & Source    & Fr-Translation   & Source    & Fr-Translation   \\ \hline \hline
\rule{0pt}{10pt}\#words           & 36,109     & 39,960         & 37,339     & 39,302         & 40,440     & 40,269         \\ \hline
\rule{0pt}{10pt}\#vocabulary      & 8,077      & 8,770          & 8,694      & 8,572          & 8,808      & 8,744          \\ \hline
\rule{0pt}{10pt}\#sentences       & 1,816      & 1,816          & 2,002      & 2,002          & 1,787      & 1,787          \\ \hline
\rule{0pt}{10pt}sentence length   & 19.9      & 22.0          & 18.6      & 19.6          & 22.6      & 22.5          \\ \hline
\rule{0pt}{10pt}\#chunks          &   --    &    1,615       &   --    &     1,579      &   --    &    1,606       \\ \hline
\rule{0pt}{10pt}\begin{tabular}[c]
{@{}l@{}}Average length\\ of chunks
\end{tabular}                     &   --    &     2.1      &   --    &     2.1      &   --    &     2.1      \\ \hline
\end{tabular}
\end{table}

These differences also act on the clustering process and the MSC method. Table~\ref{tb:st:french} details the number and the average size of clusters with at least two French sentences translated from each source language. French translations from Portuguese produced the shortest compressions (18.6 words) while compressions from Spanish had the highest compression ratio. With respect to other languages, the similarity of the sentences translated from English is lower, which leads to fewer clusters.
Summaries from Spanish have a larger proportion of compressions in the summaries than other languages.

\begin{table}[h]
\centering
\caption{\label{tb:st:french}Statistics about clusters and compressions for texts translated into French.}
\begin{tabular}{l|c|c|c}
\hline
                                                    & English  & Portuguese   & Spanish   \\\hline \hline
\rule{0pt}{10pt}\#clusters                          &    50    &     70       &     75    \\ \hline
\rule{0pt}{10pt}Average size of clusters            &    2.2   &     2.7      &     2.8   \\ \hline
\rule{0pt}{10pt}Average length of clusters          &    29.1  &     25.6     &    35.4   \\ \hline
\rule{0pt}{10pt}Average length of compressions      &    21.7  &     18.6     &    23.5   \\ \hline
\rule{0pt}{10pt}\begin{tabular}[c]{@{}l@{}}Average 
number of compressions\\  in summaries\end{tabular} &    0.7   &     0.9      &    1.3    \\ \hline
\rule{0pt}{10pt}\begin{tabular}[c]{@{}l@{}}Average 
compression rate\\  of compressions\end{tabular}    &   74.6\% &     72.6\%   &    66.4\% \\ \hline
\end{tabular}
\end{table}

We apply a similar analysis for the generation of English summaries from French, Portuguese, and Spanish source texts. As observed before for French summaries, the joint analysis still outperformed other baselines (Table~\ref{tb:clts:english}).
While CoRank obtained a large range of ROUGE scores among different languages (ROUGE-1 between 0.4602 and 0.4715), our approach obtained the best ROUGE scores for all languages with a small difference of ROUGE scores (ROUGE-1 between 0.4725 and 0.4743), which proves that our method generates more stable cross-language summaries for several languages.
Chunks spot by the syntactic pattern and the Stanford CoreNLP helped our approach to produce more informative compressions, which results in better ROUGE scores.

\begin{table}[h]
\centering
\caption{\label{tb:clts:english}ROUGE f-scores for cross-language summaries from French, Portuguese, and Spanish languages to English language.}
\begin{tabular}{l|ccc|ccc|ccc}
\hline
\multirow{2}{*}{Methods}    & \multicolumn{3}{c|}{French}  & \multicolumn{3}{c|}{Portuguese} & \multicolumn{3}{c}{Spanish}          \\
       & R-1     & R-2      & R-SU4    & R-1      & R-2      & R-SU4    & R-1           & R-2          & R-SU4    \\ \hline \hline 
\rule{0pt}{10pt}Late           &  0.4149 & 0.1030 & 0.1594 & 0.4161 & 0.1010 & 0.1576 & 0.4107 & 0.1083 & 0.1603 \\ \hline
\rule{0pt}{10pt}Early          &  0.4163 & 0.1021 & 0.1602 & 0.4135 & 0.1003 & 0.1580 & 0.4148 & 0.1132 & 0.1644 \\ \hline
\rule{0pt}{10pt}SimFusion      &  0.4179 & 0.1042 & 0.1607 & 0.4157 & 0.0999 & 0.1582 & 0.4099 & 0.1103 & 0.1616 \\ \hline
\rule{0pt}{10pt}CoRank         &  0.4645 & 0.1326 & 0.1939 & 0.4715 & 0.1415 & 0.2015 & 0.4602 & 0.1414 & 0.1966 \\ \hline
\rule{0pt}{10pt}Our approach   &  \textbf{0.4727} & 0.1375 & \textbf{0.1969} & \textbf{0.4743} & \textbf{0.1466} & \textbf{0.2047} & \textbf{0.4725} & \textbf{0.1458} & \textbf{0.2027} \\ \hline
\rule{0pt}{10pt}\begin{tabular}[c]
{@{}l@{}}Our approach\\w/o chunks
\end{tabular}                  &  0.4704 & \textbf{0.1391} & 0.1963 & 0.4731 & 0.1444 & 0.2037 & 0.4648 & 0.1393 & 0.1975 \\ \hline
\end{tabular}
\end{table}

English translations have fewer words and a smaller vocabulary (difference bigger than 1,000 words) than source texts (Table~\ref{tb:datasets:english}). These translations also have shorter sentences and a more similar vocabulary size than French translations and source texts. 
The combination of syntactic patterns and the Stanford CoreNLP led to the same characteristics of chunks in terms of numbers and sizes.

\begin{table}[h]
\centering
\caption{\label{tb:datasets:english}Statistics of datasets and their translation to English.}
\begin{tabular}{l|cc|cc|cc}
\hline
                & \multicolumn{2}{c|}{French} & \multicolumn{2}{c|}{Portuguese} & \multicolumn{2}{c}{Spanish} \\
                & Source    & En-Translation   & Source    & En-Translation   & Source    & En-Translation   \\\hline \hline
\rule{0pt}{10pt}\#words             & 41,071     & 35,929         & 37,339     & 35,244         & 40,440     & 37,066         \\\hline
\rule{0pt}{10pt}\#vocabulary        & 8,837      & 7,718          & 8,694      & 7,615          & 8,808      & 7,703          \\\hline
\rule{0pt}{10pt}\#sentences         & 2,000      & 2,000          & 2,002      & 2,002          & 1,787      & 1,787          \\\hline
\rule{0pt}{10pt}sentence length     & 20.5      & 18.0          & 18.6      & 17.6          & 22.6      & 20.7          \\\hline
\rule{0pt}{10pt}\#chunks            &   --    &     4,302      &   --    &     4,327      &   --    &     4,324      \\ \hline
\rule{0pt}{10pt}\begin{tabular}[c]
{@{}l@{}}Average length\\ of chunks
\end{tabular}                       &   --    &      2.3     &    --   &     2.3      &   --    &     2.3      \\ \hline
\end{tabular}
\end{table}

Table \ref{tb:st:english} details the clustering and the compression processes for the English translations. These translations from French source texts have more clusters because we used a smaller similarity threshold to consider two sentences as similar. English summaries from French have more compressions because of the large number of clusters.

\begin{table}[h]
\centering
\caption{\label{tb:st:english}Statistics about clusters and compressions for English translated texts.}
\begin{tabular}{l|c|c|c}
\hline
                               				   & French    & Portuguese   & Spanish    \\ \hline \hline
\rule{0pt}{10pt}\#clusters                     &    128    &     69       &    84      \\ \hline
\rule{0pt}{10pt}Average size of clusters       &    2.7    &     2.7      &    2.8     \\ \hline
\rule{0pt}{10pt}Average length of clusters     &    22.1   &     19.2     &    27.0   \\ \hline
\rule{0pt}{10pt}Average length of compressions &    16.4   &     16.3     &    21.1   \\ \hline
\rule{0pt}{10pt}\begin{tabular}[c]{@{}l@{}}
Average number of compressions\\  in summaries
\end{tabular}                                  &    2.5    &     0.9      &    1.5      \\ \hline
\rule{0pt}{10pt}\begin{tabular}[c]{@{}l@{}}
Average compression rate\\  of compressions
\end{tabular}                                  &    74.2\% &     84.9\%   &    78.1\%      \\ \hline
\end{tabular}
\end{table}

French and Portuguese source texts have almost the same number of sentences, while English and Spanish source texts have fewer sentences. Comparing the results of English and French translations, English compressions are shorter than French compressions. The use of chunks in MSC improved the results of our cross-language summaries, especially for English translations that have chunks that are more numerous and complex than French translations.

To sum up, our approach has shown to be more stable than extractive methods, thus generating more informative cross-language summaries with consistent ROUGE scores measured in several languages.

\section{Conclusion}
\label{sc:conc}

Cross-Language Text Summarization (CLTS) produces a summary in a target language from documents written in a source language. It implies a combination of the processes of automatic summarization and machine translation. Unfortunately, this combination produces errors, thereby reducing the quality of summaries.
A joint analysis allows CLTS systems to extract relevant information from source and target languages, which improves the generation of extractive cross-language summaries. Recent methods have proposed compressive and abstractive approaches for CLTS; however, these methods use frameworks or tools that are available in few languages, limiting the portability of these methods to other languages. Our Multi-Sentence Compression (MSC) approach generates informative compressions from several perspectives (translations from different languages) and achieves stable ROUGE results for all languages. In addition, our method can be easily adapted to other languages.

As future work, we plan to reduce the number of errors generated from the pipeline made of the compression and machine translation processes by developing a Neural Network method to jointly translate and compress sentences. It would also be interesting to include a neural language model to correct possible errors produced during the sentence compression process.

\section*{Acknowledgement}

This work was granted by the European Project CHISTERA-AMIS ANR-15-CHR2-0001. We also like to acknowledge the support given by the Laboratoire VERIFORM from the \'Ecole Polytechnique de Montr\'eal and her coordinator Hanifa Boucheneb.

\bibliographystyle{splncs}
\bibliography{paper}

\begin{thebibliography}{10}

\bibitem{Yao:2015:phrase_based}
Yao, J., Wan, X., Xiao, J.:
\newblock Phrase-based compressive cross-language summarization.
\newblock In: {EMNLP}. (2015)  118--127

\bibitem{Wan:2018}
Wan, X., Luo, F., Sun, X., Huang, S., Yao, J.g.:
\newblock Cross-language document summarization via extraction and ranking of
  multiple summaries.
\newblock Knowledge and Information Systems (2018)

\bibitem{elvys:2018:nldb}
Linhares~Pontes, E., Huet, S., Torres-Moreno, J.M., Linhares, A.C.:
\newblock Cross-language text summarization using sentence and multi-sentence
  compression.
\newblock In: 23rd International Conference on Natural Language \& Information
  Systems (NLDB). (2018)

\bibitem{Giannakopoulos:2011}
Giannakopoulos, G., El{-}Haj, M., Favre, B., Litvak, M., Steinberger, J.,
  Varma, V.:
\newblock {TAC2011} multiling pilot overview.
\newblock In: 4th Text Analysis Conference {TAC}. (2011)

\bibitem{Wan:2011}
Wan, X.:
\newblock Using bilingual information for cross-language document
  summarization.
\newblock In: {ACL}. (2011)  1546--1555

\bibitem{Villada:2006}
Moirón, B.V., Tiedemann, J.:
\newblock Identifying idiomatic expressions using automatic word-alignment.
\newblock In: EACL 2006 Workshop on Multiword Expressions in a Multilingual
  Context. (2006)

\bibitem{deCaseli:2010}
de~Caseli, H.M., Ramisch, C., das Gra{\c{c}}as Volpe~Nunes, M., Villavicencio,
  A.:
\newblock Alignment-based extraction of multiword expressions.
\newblock Language Resources and Evaluation \textbf{44} (2010)  59--77

\bibitem{StanfordCoreNLP14}
Manning, C., Surdeanu, M., Bauer, J., Finkel, J., Bethard, S., McClosky, D.:
\newblock The {S}tanford {CoreNLP} natural language processing toolkit.
\newblock In: 52nd Annual Meeting of the Association for Computational
  Linguistics (ACL): System Demonstrations. (2014)  55--60

\bibitem{linhares:2018:textgraph}
Linhares~Pontes, E., Huet, S., Gouveia~da Silva, T., Linhares, A.C.,
  Torres-Moreno, J.M.:
\newblock Multi-sentence compression with word vertex-labeled graphs and
  integer linear programming.
\newblock In: TextGraphs-12: the Workshop on Graph-based Methods for Natural
  Language Processing, Association for Computational Linguistics (2018)

\bibitem{filippova:2010}
Filippova, K.:
\newblock Multi-sentence compression: Finding shortest paths in word graphs.
\newblock In: COLING. (2010)  322--330

\bibitem{Wan:2010}
Wan, X., Li, H., Xiao, J.:
\newblock Cross-language document summarization based on machine translation
  quality prediction.
\newblock In: ACL. (2010)  917--926

\bibitem{boudin:2011}
Boudin, F., Huet, S., Torres{-}Moreno, J.M.:
\newblock A graph-based approach to cross-language multi-document
  summarization.
\newblock Polibits \textbf{43} (2011)  113--118

\bibitem{Zhang:2016}
Zhang, J., Zhou, Y., Zong, C.:
\newblock Abstractive cross-language summarization via translation model
  enhanced predicate argument structure fusing.
\newblock {IEEE/ACM} Trans. Audio, Speech {\&} Language Processing \textbf{24}
  (2016)  1842--1853

\bibitem{Bing:2015}
Bing, L., Li, P., Liao, Y., Lam, W., Guo, W., Passonneau, R.J.:
\newblock Abstractive multi-document summarization via phrase selection and
  merging.
\newblock In: ACL, The Association for Computer Linguistics (2015)  1587--1597

\bibitem{lin:2004}
Lin, C.Y.:
\newblock {ROUGE: A Package for Automatic Evaluation of Summaries}.
\newblock In: Workshop Text Summarization Branches Out (ACL'04). (2004)  74--81

\end{thebibliography}

\end{document}